\pdfoutput=1
\documentclass[11pt]{article}

\usepackage[]{EACL2023}

\usepackage{times}
\usepackage{latexsym}

\usepackage[T1]{fontenc}
\usepackage[utf8]{inputenc}

\usepackage{microtype}

\usepackage{inconsolata}

\usepackage{booktabs}

\usepackage{multirow}

\title{Two Approaches to Diachronic Normalization of Polish Texts}

\author{Kacper Dudzic \and
        Filip Graliński \and 
        Krzysztof Jassem \and 
        Marek Kubis \and 
        Piotr Wierzchoń \\ 
        Adam Mickiewicz University, Poznań, Poland \\      \texttt{\{firstname.lastname\}@amu.edu.pl}}
        
\begin{document}
\maketitle
\begin{abstract}
This paper discusses two approaches to the diachronic normalization of Polish texts: 
a rule-based solution that relies on a set of handcrafted patterns, and a neural normalization model based on the text-to-text transfer transformer architecture.
The training and evaluation data prepared for the task are discussed in detail, along with experiments conducted to compare the proposed normalization solutions. A quantitative and qualitative analysis is made. It is shown that at the current stage of inquiry into the problem, the rule-based solution outperforms the neural one on 3 out of 4 variants of the prepared dataset, although in practice both approaches have distinct advantages and disadvantages.
\end{abstract}

\section{Introduction}

This paper discusses two solutions to the problem of diachronic normalization, that is, the task
of determining contemporary spelling for a given historical text. Diachronic normalization may concern the writing of individual words, punctuation, hyphenation, or separation of tokens. We believe that the methods described in this paper may be useful for linguistic research on historical texts. A practical use case for our work is to facilitate full-text search in historical texts -- a query written in contemporary spelling may trigger a search for historical variants through the use of reversed-order diachronic normalization.

Similar experiments, for text normalization in a speech synthesis system
from text, were described in \cite{DBLP:journals/corr/SproatJ16}. Those authors claim that
text normalization remains one of the few tasks in the field of natural language processing where handcrafted rules may yield better results than machine learning. This is due to the following reasons:
\begin{itemize}
\item
Lack of training data; there is no economic motivation for creating
training data for text normalization -- unlike machine translation, for example, for which training data are created "naturally";
\item
Low data density of interesting cases, i.e. words that should be somehow changed -- unlike
for example, phonemic transcription, where all words are converted to a new representation;
\item
Standard methods of evaluation which do not reward trivial cases (copying of input words), thus favoring human labor.
\end{itemize}

In our experiments, we compare the results of a rule-based approach with one based on machine learning. The rule-based approach relies on a set of handcrafted rules to normalize text.
In the ML approach, we train a supervised normalization model on the basis of
a corpus of Polish books for which both historical and current spellings are available.

\section{Related work}

The first attempts at rule-based diachronic normalization
used for historical text in English were described by \citet{rayson07}
and \citet{Baron09}. Similar
studies were conducted for German \cite{Archer06}. There, context rules operated at the level of letters instead of words. The normalization rules may be derived
from corpora, as \citet{Bollman11} showed for German. Diachronic normalization
may be also performed using a noisy channel
model, as described by \citet{oravecz10} using the example
of Old Hungarian texts. Research on diachronic
normalization has also been conducted for Portugese \cite{reynaert12}, Swedish \cite{petterson12}, Slovene \cite{Scherrer13}, Spanish \cite{porta13},
 and Basque \cite{Etxeberia16}.

\citet{bollmann-2019-large} surveys historical spelling normalization methods for eight languages. 
He reports word-level accuracy for the evaluated systems.
He claims that using CER is not justified, because it strongly correlates with WER for systems showing reasonable accuracy.
\citet{bollmann-sogaard-2016-improving} use bi-directional LSTMs and multi-task learning to normalize texts in Early New High German. 
Their dataset consists of 44 texts from the Aselm corpus.
he model presented is evaluated with respect to word-level accuracy.
\citet{robertson-goldwater-2018-evaluating} discuss the problem of evaluating historical text normalization systems.
They emphasize the necessity of reporting accuracy for unobserved tokens and recommend confronting the normalization systems with a simple baseline that memorizes training samples.

\citet{jassem17pros,jassem18automatic} present an automatic method for diachronic normalization of Polish texts.
The proposed method uses a formal language to model diachronic changes.
\citet{GralinskiMining2020} introduce a method for finding spelling variants in a diachronic corpus using word2vec.

\section{Data}

Training and evaluation of a diachronic normalizer requires a corpus of texts that preserve
historical spelling along with their contemporized counterparts.
As our aim is the normalization of Polish prose, we decided to collect texts for our corpus
from two sources.
Texts that preserve historical spelling were drawn from the Polish edition of the Wikisource
project \cite{wikisource23}, which provides proof-read transcriptions of printed books that have fallen into the public domain,
encoded in the MediaWiki format.
For contemporized texts, we used Wolne Lektury \cite{wolnelektury23}, a digital library that aims to deliver
new editions of school readers, free of charge.
Although both sources encompass a wide variety of texts, ranging from poems and works of philosophy to dictionaries and historical
documents, we narrowed our attention to novels,
to facilitate the process of matching the original texts from Wikisource to their contemporized versions
in Wolne Lektury with the use of metadata information available for novels in both sources.
We initially sourced 308 novels from Wikisource and 279 from Wolne Lektury.

\subsection{Preprocessing}

All of the texts then underwent preprocessing.
First, we split the texts into paragraphs, with the use of markup information preserved in XML files sourced from Wolne Lektury, and MediaWiki content collected from Wikisource.
Next, regular expressions were used to remove leftover markup information, such as in-text metadata, formatting, or HTML tags, and to normalize some atypical characters. Accordingly, diacritical marks were removed from letter characters not belonging to the Polish alphabet, and non-ASCII variants of standard letter characters of the Latin alphabet were replaced by their ASCII counterparts. Finally, the same method was used to remove dialogue-specific text formatting and punctuation in paragraphs consisting of dialogue utterances, such as quotation dashes or character cues.

\subsection{Alignment}

To create aligned paragraph data, we first automatically matched all editions of novels existing across both data sources using fuzzy information similarity for author and title metadata. We then narrowed the matches to those that contained at least one edition in each of the sources.

Next, for each match of all editions of a novel, the oldest edition from Wikisource and the most recent edition from WolneLektury were identified using metadata information. Subsequently, the text paragraphs of both editions were extracted and aligned using the Hunalign tool, version 1.1 \cite{varga2005parallel}. Specifically, it was used to automatically create paragraph pairs consisting of a given text fragment with historical spelling from the oldest edition of a novel and the same text fragment but with contemporized spelling found in its newer edition, optionally automatically joining or splitting paragraphs where it was applicable. The paragraph alignment quality metric returned by Hunalign was consulted to provide additional filtering. The average alignment quality score across the entire text contents for each edition pair was used to identify and discard very low-scoring edition pairs, which turned out to be Polish translations of foreign novels made by different translators. In turn, per-paragraph alignment quality scores below 1.0 were used as an indicator to discard singular misaligned paragraphs.

\subsection{Dataset creation}
After completing all of the above steps and performing deduplication at the very end, we obtained a final corpus of 248,645 paragraph pairs originating from 87 eligible pairs of matched novel editions. Four dataset variants were created with this as the basis. All variants involve a training and test split, but they differ in the following two respects:

\begin{description}
\item[Pruning] was either applied or not. \emph{Pruned} versions of the dataset are reduced in size by removing samples in which the paragraphs of the pair are identical. Applying pruning leads to a 64.83\% decrease in the number of samples, a 47.34\% decrease in the number of words, and a 47.23\% decrease in the number of characters.
\item[Separation] of novels prior to the train/test split was either performed or not.
In \emph{separated} variants of the dataset, train and test sets are created from separate pools of novels with no overlap, so that all paragraphs from a given novel are contained in only one of the sets.
Four novels were sampled from each of the quartiles determined with respect to the number of paragraphs contained in the corpus, to guarantee that each data subset contained a balanced volume of text.
In the case of \emph{non-separated} variants, the paragraphs are randomly sampled from the entire set of novels following the standard $80\%/20\%$ sampling ratio for train/test splits.

\end{description}

\begin{table*}
    \centering
    \begin{tabular}{llrrrr}
    \toprule
        \multirow{2}{*}{\textbf{Pruning}} & \multirow{2}{*}{\textbf{Separation}} & \multicolumn{2}{c}{\textbf{Split samples}} & \multirow{2}{*}{\textbf{Characters}} & \multirow{2}{*}{\textbf{Words}} \\
    \cmidrule(lr){3-4}
        & & \footnotesize{Train} & \footnotesize{Test} & & \\
    \midrule
        No  & No  & 198,916 & 49,729 & 92,306,901 & 14,438,223 \\
        Yes & No  & 69,952  & 17,488 & 48,710,393 & 7,603,573  \\
        No  & Yes & 199,004 & 49,641 & 92,306,901 & 14,438,223 \\
        Yes & Yes & 63,921  & 23,519 & 48,710,393 & 7,603,573  \\
    \bottomrule
    \end{tabular}
    \caption{Dataset statistics}
    \label{tab:dataset}
\end{table*}

\begin{table*}
    \centering
    \begin{tabular}{lllrrrr}
    \toprule
        \textbf{Method} & \textbf{Pruning} & \textbf{Separation} & \textbf{CER} & \textbf{WER} \\
    \midrule
        Transducers & No  & No   & \textbf{0.0164} & \textbf{0.0466} \\
        Neural & No  & No   & 0.0488 & 0.0654 \\
    \midrule
        Transducers & Yes & No   & \textbf{0.0319} & \textbf{0.0827} \\
        Neural & Yes & No   & 0.0728 & 0.1011 \\
    \midrule
        Transducers & No  & Yes  & \textbf{0.0182} & \textbf{0.0560} \\
        Neural & No  & Yes  & 0.0632 & 0.0932 \\
    \midrule
        Transducers & Yes & Yes  & \textbf{0.0281} & 0.0844 \\
        Neural & Yes & Yes  & 0.0398 & \textbf{0.0737} \\
    \bottomrule
    \end{tabular}
    \caption{Evaluation results}
    \label{tab:results}
\end{table*}

\section{Experiments}
\subsection{Rule-based model}
\label{sec:rule-based-model}
Our first solution to the problem of diachronic normalization relies on a set of deterministic rules.
Henceforth, we will refer to this solution as \textit{Transducers}.
The rules were handcrafted initially and then adjusted
semi-automatically. 
They were created mostly based on the expert literature describing changes in the Polish spelling system and by looking at a list of similar words having close embeddings. For most of the work on the rules, datasets for supervised learning were not consulted.
Originally, the rules were written using the Thrax
language \cite{tai11} for defining transducer grammars,
but more recently have been rewritten into a Java code base with
normalization rules encoded using regular expressions.
For instance, the rule:

\begin{verbatim}
Rule(
    "([cs]|(?:\\A|(?<![cdsr]))z)
     y([aąeęiou])",
    "$1j$2")
\end{verbatim}

\noindent handles normalization of \textit{y} into \textit{i} in some
circumstances (e.g. \textit{decyzya} into \textit{decyzja}).
The decision to switch to Java was motivated by the fact that such a
module can be easily incorporated, as a plugin, into Java-based
open-source search engines (Lucene and Solr).
When writing the rules, a conservative approach was taken: a rule was
added only when the probability of unwanted changes to texts was very
low.
Apart from regular expressions, the rule-based solution uses a
dictionary of transformations for specific words and dictionaries of
exceptions, based on the ideas outlined in \cite{GralinskiMining2020}.
The \textit{Transducers} module also handles some OCR errors, but the coverage
is rather low (only high-precision rules were applied).

Some further examples of the rules used are included in appendix ~\ref{sec:appendix}.

\subsection{Neural normalization models}
\label{sec:neural-models}
Diachronic normalization is an example of a language processing task that accepts text at the input and returns text at the output.
Therefore, we decided to use the
text-to-text transfer transformer architecture \cite[T5,][]{raffel20exploring} as a basis for
our supervised normalization models.
Initial weights were taken from the pre-trained plT5 model \cite{chrabrowa-etal-2022-evaluation},
an encoder-decoder model that follows the T5 architecture. The
plT5 model was initialized from its multilingual counterpart \cite[mT5,][]{xue-etal-2021-mt5} and further trained on Polish language corpora. It achieves better performance than mT5 on Polish language benchmark tasks with a smaller number of parameters.
For our experiments, we used the largest variant of this model available at Hugging Face.\footnote{\url{https://huggingface.co/allegro/plt5-large}}

We finetuned four neural diachronic normalization models, with one model for each variant of our
dataset.
The models were trained for three epochs, using Adam as the optimizer and a learning rate of 5e-05 with a linear scheduler. The batch size was kept at 1 due to the memory limitations of the GPU used for the experiments. Maximum input and output sequence token lengths followed the T5 model family's default of 512. Longer input sequences were split into chunks of maximum length, processed separately, and then joined.

Table~\ref{tab:results} reports the results of the evaluation of the neural normalization models,
and compares them with the rule-based model. One may observe that the rule-based model is
a strong baseline for the task, outperforming the neural models with respect to
character error rate (CER) and word error rate (WER). However, the supervised model surpasses the
rule-based solution in the case of a test set that consists of a separate set of novels (\emph{Separation=Yes})
and excludes samples that should remain unmodified in the
normalization process (\emph{Pruning=Yes}).

\section{Discussion}

After performing a qualitative analysis of the results obtained using rule-based and neural normalization models, we observed for the neural networks: (1) flexibility in context interpretation, i.e., the ability to adapt to various contexts and understand linguistic nuances; (2) recognition of irregular patterns, i.e., the ability to identify and process non-standard and complex language forms; (3) context-based changes, i.e., considering a broad context, which can lead to changes that go beyond simple spelling rules. 
On the other hand, for rule-based normalization, it was noted that: (1) relying on specific, defined rules, i.e., focusing on the strict application of established spelling rules, \emph{Transducers} are less flexible in interpretation, meaning that they have limited abilities to cope with irregularities and linguistic nuances; (2) \emph{Transducers} follow a literal interpretation of rules, which may not take into account the full context. 
The neural approach effectively normalizes examples of former single-word spelling, especially for conjunctions: \emph{przyczem} $\to$ \emph{przy czym} (Eng. \emph{at the same time}), \emph{poczem} $\to$ \emph{po czym} (Eng. \emph{thereafter}), \emph{napewno} $\to$ \emph{na pewno} (Eng. \emph{certainly}), \emph{niema} $\to$ \emph{nie ma} (Eng. \emph{there is no}). 
The rule-based approach, in turn, aptly converts regular orthographic phenomena: \emph{egzystencya} $\to$ \emph{egzystencja} (Eng. \emph{existence}), \emph{jenerał} $\to$ \emph{generał} (Eng. \emph{general}), \emph{teorya} $\to$ \emph{teoria} (Eng. \emph{theory}). It also accurately transforms proper nouns: \emph{Anglja} $\to$ \emph{Anglia}, \emph{Marjetka} $\to$ \emph{Marietka}. 
The spelling changes – from \emph{egzystencya} to \emph{egzystencja}, \emph{Anglja} to \emph{Anglia}, etc. – were part of the Polish orthographic reform of 1936. 
This reform was aimed at simplifying and standardizing the Polish language's spelling. It introduced several changes, including the replacement of the letter 'y' with 'j' or 'i' in certain contexts, and the introduction of the letter 'j' in place of 'i' in some cases to better reflect pronunciation. 
This reform significantly influenced the modern Polish language, aligning it more closely with its phonetics.

\section{Conclusion}
This paper has discussed two approaches to the diachronic normalization of Polish texts. We presented \emph{Transducers}, a rule-based solution that relies on a set of deterministic, handcrafted rules, and a family of neural normalization models based on a text-to-text transfer transformer architecture. The experiments that we conducted showed that the rule-based approach is effective in the diachronic normalization task. However, the neural model surpassed the rule-based solution in the case of a test set that consists of a separate set of novels and excludes samples that should remain unchanged in the normalization process.

As the presented research is preliminary in nature, there are several promising directions to explore, which we are committed to doing in the near future. Among other ideas, we want to test the performance of hybrid solutions combining both approaches in distinct ways. We are also considering testing different model architectures and conducting further work on improving the quality of the training data used for the neural approach, as we believe it has the potential to eventually surpass the rule-based solution in most typical scenarios.

\section*{Limitations}
We restrict our attention to the diachronic normalization of Polish texts.
Generalizing the proposed methods to new languages will require, firstly, a new, handcrafted set of normalization rules being developed for the rule-based model presented in section~\ref{sec:rule-based-model}; and secondly, a parallel corpus of texts that encompass both historical and current spelling, for the neural normalization models discussed in section~\ref{sec:neural-models}.

\section*{Acknowledgments}
The diachronic normalization methods presented in this work have been developed as part of the ,,Tools for text normalization and diachronic analysis'' module of the Dariah.lab\footnote{\url{https://lab.dariah.pl/en/project/about-project/}} infrastructure hosted at the Faculty of Mathematics and Computer Science of Adam Mickiewicz University in Poznań, Poland. Construction of Dariah.lab was the aim of the project: \textit{Digital Research Infrastructure for the Arts and Humanities DARIAH-PL}, funded by the Intelligent Development Operational Programme, Polish National Centre for Research and Development, ID: POIR.04.02.00-00-D006/20.
% EACL 2023 requires all submissions to have a section titled ``Limitations'', for discussing the limitations of the paper as a complement to the discussion of strengths in the main text. This section should occur after the conclusion, but before the references. It will not count towards the page limit.

% The discussion of limitations is mandatory. Papers without a limitation section will be desk-rejected without review.
% ARR-reviewed papers that did not include ``Limitations'' section in their prior submission, should submit a PDF with such a section together with their EACL 2023 submission.

% While we are open to different types of limitations, just mentioning that a set of results have been shown for English only probably does not reflect what we expect.
% Mentioning that the method works mostly for languages with limited morphology, like English, is a much better alternative.
% In addition, limitations such as low scalability to long text, the requirement of large GPU resources, or other things that inspire crucial further investigation are welcome.

\bibliography{anthology,custom}
\bibliographystyle{acl_natbib}

\appendix

\section{Appendix}
\label{sec:appendix}

Here we present further examples of rules used in the rule-based diachronic normalization solution.

\begin{verbatim}
Rule("izk", "isk")
Rule("yja\\b", "ja")
Rule("(le|ó)dz\\Z", "$1c")
Rule("\\Aanti-?", "anty")
Rule("iemi\\Z", "imi")
Rule("emi\\Z", "ymi")
Rule(
    "(ąc|owan|yjn|owat|jsz|tyczn|logiczn)
    em\\Z",
    "$1ym")
Rule(
    "([dfglmnprt])[jy]([aąeęiou])",
    "$1i$2")
\end{verbatim}

\end{document}